%% file: main.tex
\documentclass[10pt,twocolumn,letterpaper]{article}

\usepackage{cvpr}              
\usepackage{multirow}
\usepackage{algorithm}
\usepackage{algorithmic}

\input{preamble}

\definecolor{cvprblue}{rgb}{0.21,0.49,0.74}
\usepackage[pagebackref,breaklinks,colorlinks,citecolor=cvprblue]{hyperref}

\def\paperID{16952} 
\def\confName{CVPR}
\def\confYear{2024}

\title{Initialization Matters for Adversarial Transfer Learning}

\author{\textbf{Andong Hua$^1$\quad Jindong Gu$^2$\quad Zhiyu Xue$^1$\quad Nicholas Carlini$^3$\quad Eric Wong$^4$\quad Yao Qin$^{1,3}$}\\
$^1$University of California, Santa Barbara \quad $^2$University of Oxford\\
$^3$Google \quad 
$^4$University of Pennsylvania
}

\begin{document}
\maketitle
\input{sec/0_abstract}
\input{sec/1_intro}

\input{sec/2_related_works}

\input{sec/3_PEFT}

\input{sec/4_pretrain}
\input{sec/5_methods_arxiv_1}

\input{sec/6_ablations}
\input{sec/7_conclusion_arxiv_1}
{
    \small
    \bibliographystyle{ieeenat_fullname}
    \bibliography{main}
}

\input{sec/X_suppl_arxiv}

\end{document}

%% file: preamble.tex
\usepackage[dvipsnames]{xcolor}


%% file: sec/0_abstract.tex
\begin{abstract}

With the prevalence of the Pretraining-Finetuning paradigm in transfer learning, the robustness of downstream tasks has become a critical concern. In this work, we delve into adversarial robustness in transfer learning and reveal the critical role of initialization, including both the pretrained model and the linear head. First, we discover the necessity of an adversarially robust pretrained model. Specifically, we reveal that with a standard pretrained model, Parameter-Efficient Finetuning~(PEFT) methods either fail to be adversarially robust or continue to exhibit significantly degraded adversarial robustness on downstream tasks, even with adversarial training during finetuning. Leveraging a robust pretrained model, surprisingly, we observe that a simple linear probing can outperform full finetuning and other PEFT methods with random initialization on certain datasets. We further identify that linear probing excels in preserving robustness from the robust pretraining. Based on this, we propose Robust Linear Initialization~(RoLI) for adversarial finetuning, which initializes the linear head with the weights obtained by adversarial linear probing to maximally inherit the robustness from pretraining. Across five different image classification datasets, we demonstrate the effectiveness of RoLI and achieve new state-of-the-art results. Our code is available at \url{https://github.com/DongXzz/RoLI}.

\end{abstract}

%% file: sec/1_intro.tex
\section{Introduction}
\label{sec:intro}

\begin{figure}[t]
  \centering
   \includegraphics[width=0.9\linewidth]{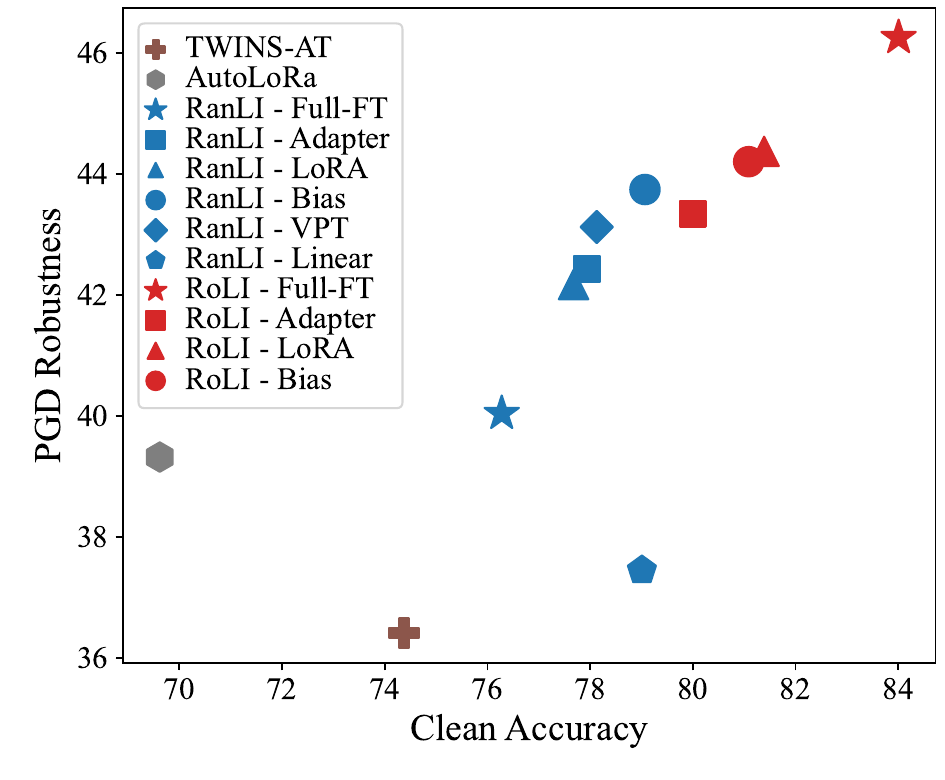}
   \vspace{-4mm}
   \caption{\textbf{Robust Linear Initialization~(RoLI), significantly improves adversarial robustness.} RoLI, denoted in red, achieves an average 3.88\% increase in clean accuracy and a 2.44\% increase in robust accuracy compared to Random Linear Initialization (RanLI) of the linear head during adversarial finetuning, averaged across five downstream datasets. Our best-performing RoLI - Full-FT, which represents adversarial full finetuning with robust linear initialization, achieves a new state-of-the-art performance. We include six popular finetuning methods with Swin Transformer~\cite{swin} and two existing state-of-the-art techniques for adversarial transfer learning: TWINS-AT~\cite{twins} and AutoLoRA~\cite{autolora}.}
   \label{fig:intro}
   \vspace{-5mm}

\end{figure}

With the advancement of large-scale deep learning models, the Pretraining-Finetuning paradigm takes a more dominant role compared to training from scratch across various tasks~\cite{foundationmodel}, including computer vision~\cite{vit,mae,sam,swin}, natural language processing~\cite{attn,bert,raffel2020exploring}, and speech recognition~\cite{speech}. Under the paradigm of Pretraining-Finetuing, advanced parameter-efficient finetuning~(PEFT) methods~\cite{lora,vpt,adapter1,ss,glora,noah} have emerged. Compared to full finetuning, PEFT methods either introduce small modules into a fixed pretrained model or optimize only part of the pretrained model, demonstrating exceptional performance while keeping storage usage low.

\begin{figure*}[t]
    \centering
    \includegraphics[width=1\textwidth]{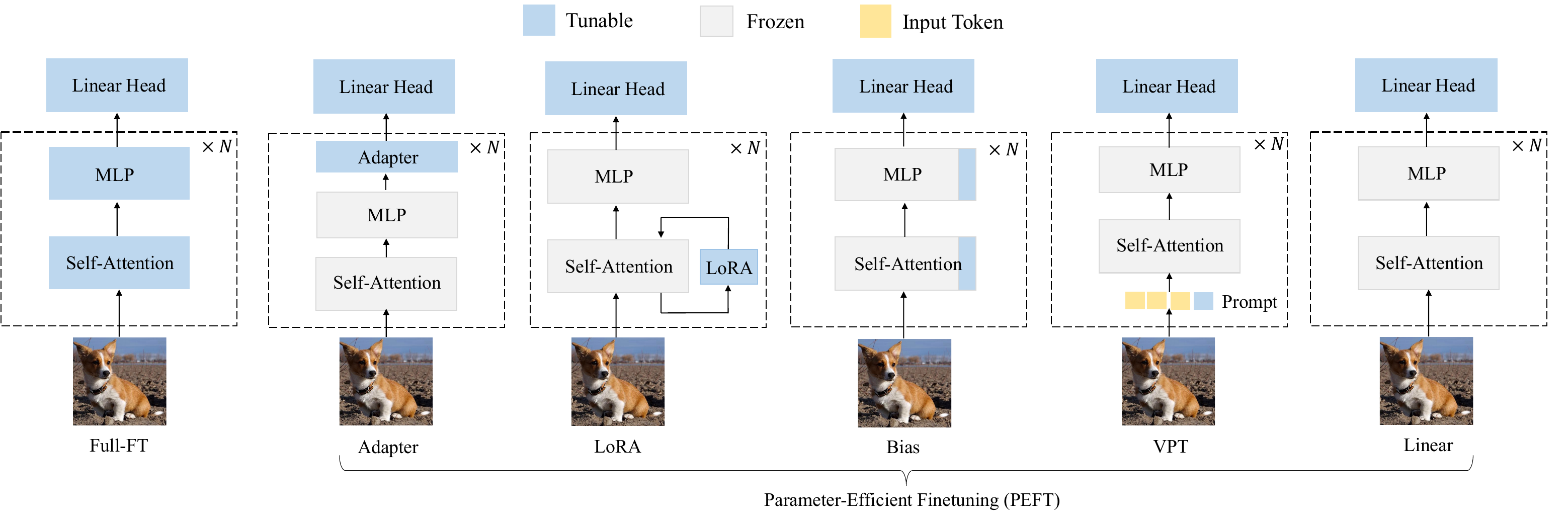}
    \caption{Illustration of six different finetuning techniques, arranged in descending order according to the number of tunable parameters.}
    \label{fig:PEFTs}
    \vspace{-5mm}
\end{figure*}
While the success of transfer learning is evident, the robustness of downstream models has become a critical concern in real-world applications. For example, as suggested by previous studies, full finetuning will distort the pretrained features, leading to compromised robustness in terms of image corruptions and out-of-distribution~(OOD) performance~\cite{yamada2022does,kumar2022fine}. In addition, adversarial robustness poses another significant challenge to real-world deployment, as a non-robust model demonstrates nearly zero accuracy under adversarial attacks~\cite{shao2022adversarial}. Many approaches have been proposed to enhance adversarial robustness in transfer learning, including improving the robustness of pretraining through self-supervised techniques~\cite{chen2020adversarial,jiang2020robust} and preserving robustness from pretrained models during finetuning~\cite{twins,autolora}. However, most previous works~\cite{autolora,twins} apply full finetuning for adversarial transfer learning, overlooking the significance of different finetuning methods.

In this work, we comprehensively study six popular finetuning methods in adversarial transfer learning, as shown in \cref{fig:PEFTs}.
We discover that initialization, which includes 1) the backbone initialization (i.e., a pretrained model), and 2) the initialization of the linear head that adapts features to the target domain, plays a critical role in adversarial transfer learning. We start by comparing the performance of adversarial finetuning when initialized with either a robust or a standard pretrained model. Surprisingly, all PEFT methods fail or exhibit significantly inferior performance when initialized with a standard pretrained model compared to being initialized with a robust pretrained model.
Furthermore, we observe that a large-scale pretrained model, such as CLIP~\cite{clip}, cannot alleviate this phenomenon, highlighting the critical role of robust pretraining in adversarial transfer learning.

Given a robust pretrained model, surprisingly, we discover that adversarial linear probing outperforms other methods on certain datasets, e.g., Caltech256~\cite{caltech} and Stanford Dogs~\cite{stanforddogs}. We further demonstrate that this is mainly because linear probing excels in inheriting robustness from pretraining. Moreover, we identify the ability to adapt features to the specific target domain as a key factor influencing linear probing's success, supported by a strong correlation between transferred accuracy and robustness. Based on these insights, we introduce Robust Linear Initialization (RoLI), which initializes the model's linear head with the weights obtained from adversarial linear probing. With RoLI followed by adversarial finetuning methods, we maximize the inherited robustness along with a strong feature adaptation ability, resulting in improved performance. In summary, our contributions are as follows:

\begin{itemize}
    \item We comprehensively study six popular finetuning techniques for adversarial transfer learning. We discover that PEFT methods fail or exhibit significantly inferior performance when initialized with a standard pretrained model, even with adversarial finetuning on downstream data. 
    \item We demonstrate that adversarial linear probing excels in preserving robustness from a robust pretrained model. Building upon this insight, we propose Robust Linear Initialization~(RoLI) for adversarial finetuning to maximally inherit robustness from pretraining and effectively adapt features through adversarial finetuning.
    \item We demonstrate the effectiveness of RoLI across five downstream datasets. On average, RoLI 
    improves the clean and robust accuracy by 3.88\% and 2.44\% compared with random initialization. This establishes a new state-of-the-art benchmark for adversarial transfer learning.
\end{itemize}

%% file: sec/2_related_works.tex
\section{Related Works}
\label{sec:related_works}

\paragraph{Finetuning in Transfer Learning.}
Transfer learning aims to finetune a pretrained model on downstream tasks to gain better performance~\cite{yosinski2014transferable,imagenetTransferBetter,zamir2018taskonomy}. Linear probing and full finetuning are often applied in the finetuning process. Given the increasing size of pretrained models, various Parameter Efficient Finetuning~(PEFT) methods have been proposed. PEFT methods can effectively reduce the finetuning cost and alleviate overfitting since only a small number of parameters are updated~\cite{adapter1,lora}. Specifically, Bias~\cite{bias1,bias2} only updates the bias term, while Scaling \& Shifting~\cite{ss} performs the linear transformation to adapt the features. Different from both of them, Adapter~\cite{adapter1,adapter2,adapter3}, LoRA~\cite{lora}, and VPT~\cite{vpt} introduce extra learnable modules into the pretrained models in the finetuning stage. Although there are various finetuning methods proposed to improve the accuracy, it is still unclear what matters to the robustness on the downstream tasks. \cite{kumar2021LPFT} points out that feature distortion in finetuning hurts the out-of-distribution robustness on downstream tasks, while we discover initialization plays a critical role in adversarial robustness on downstream tasks.

\paragraph{Adversarial Robustness in Transfer Learning.}
The research on adversarial robustness in transfer learning has been studied from two perspectives, namely, robust pretraining and robust finetuning. Concretely, robust pertaining aims to achieve adversarial robustness in the pertaining stage and finetune pretrained models on the downstream tasks without adversarial training~\cite{chen2020adversarial,jiang2020robust,cartl,artl}. Besides, \cite{salman2020adversarially} demonstrates that the adversarially pertaining also benefits standard performance on downstream tasks. In contrast, robust finetuning~\cite{twins,autolora} focuses on preserving the robustness of the pretrained model during the finetuning stage. For example, TWINS~\cite{twins} incorporates a dual batch normalization in the model to keep the statistics of pretrained and finetuned datasets separately. And AutoLoRA~\cite{autolora} introduces a low-rank~(LoRA) branch to disentangle clean and adversarial objectives. Existing works design various strategies to preserve the robustness from the pretrained model, but they simply apply full finetuning on downstream tasks. In this work, we study six different finetuning methods and introduce Robust Linear Initialization (RoLI) to enhance the adversarial robustness for different finetuning methods.

%% file: sec/3_PEFT.tex
\section{Background}
\label{sec:background}

In this section, we provide an overview of existing finetuning techniques, including full finetuning as well as five different PEFT methods, as illustrated in \cref{fig:PEFTs}. Additionally, we introduce adversarial finetuning which integrates adversarial training during the finetuning stage. 

\subsection{Existing Finetuning Methods}
\label{subsec:PEFTs}

In transfer learning, finetuning adapts the features from the pretrained domain to the target domain. In this section, we introduce six finetuning methods in descending order according to the number of tunable parameters.

\noindent\textbf{Full Finetune~(Full-FT)}: We initialize the model from a pretrained model and finetune all its parameters on the downstream tasks.

\noindent \textbf{Adapter}~\cite{adapter1,adapter2,adapter3}:  
Adapter introduces a module after the MLP block in every layer. The adapter module consists of a down-sampling layer and an up-sampling layer, with a non-linear activation. During finetuning, we will update the adapter modules as well as the linear head.

\noindent \textbf{Low-Rank Adaptation~(LoRA)}~\cite{lora}: LoRA proposes to learn a residual from pretrained weight represented by two low-rank metrics. Specifically, we apply the LoRA branch on the query and value projection in the self-attention block. During finetuning, the pretrained model remains frozen while updates are applied to the LoRA branch and head.

\noindent \textbf{Bias}~\cite{bias1,bias2}: We update the bias term and keep other parameters unchanged. Additionally, We also update the classification head~(including weights and bias).

\noindent \textbf{Visual Prompt~Tuning (VPT)}~\cite{vpt}: VPT appends additional learnable tokens (embeddings), called prompts, into the input space of each attention layer in vision transformers. Specifically, we use the structure of VPT-Deep, where every layer introduces a fixed number of trainable prompts independently. During the finetuning stage, we will tune both the prompts and the linear head on downstream tasks while freezing the entire pretrained backbone. We refer readers to~\cite{vpt} for more details.

\noindent \textbf{Linear Probing~(Linear)}: We only finetune the classification head on the downstream tasks.

\subsection{Adversarial Finetuning}
Adversarial training corresponds to a min-max optimization problem~\cite{madry2018adver}. In the context of adversarial finetuning, we can formulate the optimization problem as follows:
\begin{equation}
    \min _{\hat{\theta}} \mathbb{E}_{(x, y) \sim D}\left[\max _{\|\delta\|_\infty \leq \varepsilon} \mathcal{L}(x+\delta, y ; \theta \cup \hat{\theta} )\right]
\end{equation}
where $\theta$ and $\hat{\theta}$ represent the frozen parameters and tunable parameters in model parameters respectively. For example, in adversarial full finetuning, $\hat{\theta}$ stands for all the model parameters as we update all of them during the finetuning stage. In contrast, in adversarial VPT, $\hat{\theta}$ denotes the extra prompts and linear head while $\theta$ represents the frozen parameters in the backbone architecture. In addition, $\delta$ represents adversarial perturbation, whose $\lVert\cdot\rVert_\infty$ is bounded by $\varepsilon$. During adversarial finetuning, we use the pretrained model together with additional modules introduced by PEFT methods to generate adversarial examples, while only updating the tunable parameters $\hat{\theta}$ for each method. If not otherwise specified, we use PGD-7 with ${\varepsilon} = 8/255$ and step size ${\alpha}= 2/255$ to generate adversarial attacks during adversarial finetuning.

%% file: sec/4_pretrain.tex
\section{Robust Pretrained Model Matters}
\label{sec:pretrain}

\begin{figure*}[t]
  \centering
   \includegraphics[width=1\linewidth]{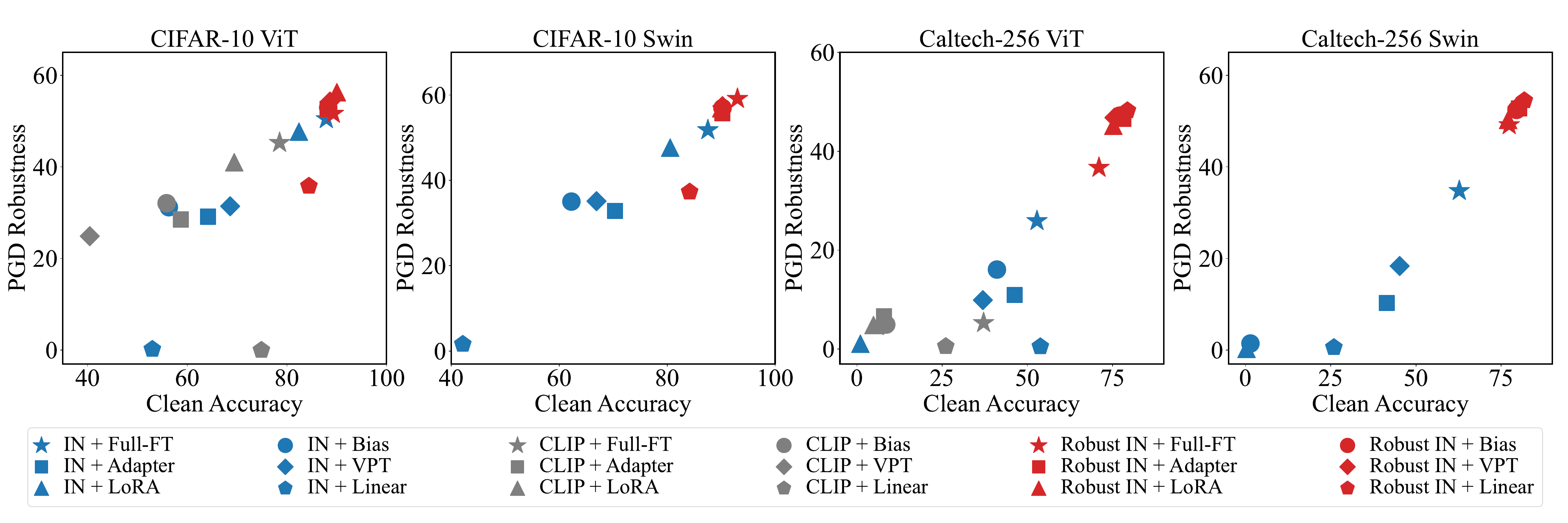}
   \caption{\textbf{PEFT methods fail or exhibit significantly degraded performance with a standard pretrained model~(in blue and gray).} Models finetuned from a robust pretrained model~(in red) exhibit high accuracy and robustness, as they are consistently positioned in the top-right corner. Among six finetuning techniques, Full-FT outperforms others when starting with a standard pretrained model; however, it falls short of other methods when starting with a robust pretrained model. The specific numerical results are provided in the supplementary.}
   \label{fig:pretrain}
   \vspace{-5mm}
\end{figure*}

Previous research~\cite{twins,artl} suggests that although standard pretraining results in performance inferior to that of adversarially robust pretraining, it can still yield an adversarially robust model with adversarial finetuning. In this section, our goal is to investigate the significance of pretraining in PEFT methods. Specifically, we seek to address the question: \textit{Is an adversarially robust model necessary for PEFT methods?}

To answer this question, we initialize model parameters using two different pretrained models. One is a standard pretrained model on ImageNet-1k~\cite{imagenet} while the other is adversarially pretrained with PGD~\cite{madry2018adver} attacks. These pretrained models are off-the-shelf and their specifics are outlined in the supplementary material. Following initialization, we perform adversarial finetuning on downstream tasks. As shown in \cref{fig:pretrain}, we evaluate six adversarial finetuning techniques across two model architectures~(ViT-B~\cite{vit} and Swin-B~\cite{swin}) on the CIFAR10~\cite{cifar} and Caltech256~\cite{caltech}. The adversarial robustness is measured with PGD-10 bounded with ${\varepsilon} = 8/255$. 

First, we find that initializing the model with an adversarially robust pretrained model can significantly improve adversarial robustness on downstream tasks. This consistently holds across all the experimental settings we have evaluated, e.g., 6 adversarial finetuning techniques across 2 model architectures on 2 different datasets. More surprisingly, we discover that a robust pretrained model is especially important for PEFT methods, as they struggle to achieve satisfactory adversarial robustness when initialized with a standard pretrained model. For example, all the PEFT methods fail to achieve adversarial robustness higher than 20\% on the Caltech256 dataset. In addition, adversarial linear probing can only achieve $<5\%$ adversarial robustness on both datasets when initiated with a standard pretrained model. This strongly suggests that adversarial PEFT can not sufficiently infuse adversarial robustness in the finetuning stage if the model starts with a non-robust pretrained model. Finally, we observe that Full-FT consistently outperforms other finetuning methods with a non-robust pretraining, indicating the preference for using full finetuning in the absence of robust pretraining.
\vspace{-3.5mm}

\paragraph{Does Pretraining on a Larger Dataset Help Adversarial Finetuning?} Due to the high computational cost of adversarial pretraining, we aim to investigate whether a model pretrained on a larger dataset could substitute for the necessity of an adversarially robust pretrained model. To this end, we initialize the model with an off-the-shelf large-scale pretrained model, CLIP~\cite{clip}, and perform the same adversarial finetuning techniques on downstream tasks. \cref{fig:pretrain} demonstrates that a standard pretrained model on a larger dataset, like CLIP, does not yield higher robustness compared to the non-robust ImageNet-1k pretrained model. It notably lags far behind the robustness achieved by a smaller-scale robust pretrained model, such as one trained on ImageNet-1k. This observation implies that opting for a smaller yet robust pretrained model proves to be a more effective strategy for attaining adversarial robustness in downstream tasks compared to using a larger, non-robust pretrained model.


\begin{figure*}[t]
  \centering
   \includegraphics[width=1\linewidth]{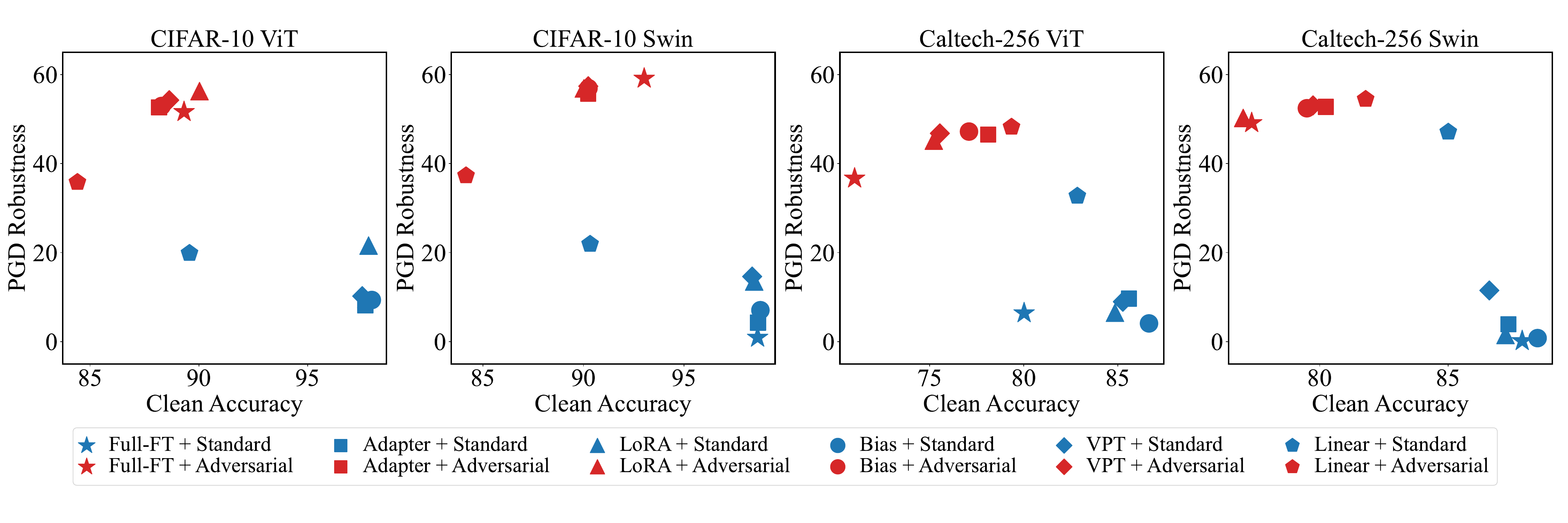}
   \vspace{-7mm}
    \caption{\textbf{Linear probing with adversarial finetuning surpasses other methods on Caltech256.} With standard finetuning, the PEFT methods exhibit inherited robustness (Robustness $>$ 0) from the pretraining. Notably, linear probing surpasses the other methods by a substantial margin in robustness while maintaining similar clean data accuracy on Caltech256. Additionally, adversarial training during finetuning is effective in enhancing all methods' robustness.}
   \label{fig:confounder}
   \vspace{-5mm}

\end{figure*}

%% file: sec/5_methods_arxiv_1.tex
\section{Initialization Matters For Finetuning}
\label{sec:methods}
In the previous section, we understand the necessity of initializing a model with an adversarially robust pretrained model. We now ask: \emph{given a robust pretrained model, what matters for adversarial finetuning?}


\begin{table}[]
\small
\setlength\tabcolsep{0.3cm}
\centering
\begin{tabular}{l|cc|cc}
\toprule
\multirow{2}{*}{Methods} & \multicolumn{2}{c}{ViT} & \multicolumn{2}{|c}{Swin} \\
                         & Clean      & PGD        & Clean       & PGD        \\
\midrule
RanLI - Full-FT          & 80.17      & 44.15      & 85.19       & 54.13      \\
RanLI - Adapter          & 83.14      & 49.57      & 85.24       & 54.22      \\
RanLI - LoRa             & 82.63      & 50.72      & 83.52       & 53.57     \\
RanLI - Bias             & 82.69      & 50.07      & 84.88       & 54.69      \\
RanLI - VPT              & 82.09      & 50.52      & 84.99       & 55.23      \\
RanLI - Linear           & 81.89      & 42.02      & 82.98       & 45.89      \\
\bottomrule
\end{tabular}
\caption{\textbf{PEFT methods, excluding Linear, demonstrate strong performance on average across CIFAR10 and Caltech256 datasets.} With adversarial finetuning, PEFT methods outperform Full-FT and Linear with RanLI. Full results are available in \cref{tab:numerical_confounder}}
\label{tab:RanLI}
\vspace{-5mm}

\end{table}

\subsection{Random Linear Initialization}\label{sec:ranli}

To answer this question, we first evaluate the adversarial robustness of six adversarial finetuning techniques. To adapt the robust pretrained model to the downstream tasks, we initialize the linear head with Random Linear Initialization (RanLI) following~\cite{artl}. We perform adversarial finetuning across two model architectures, e.g., ViT and Swin, on the CIFAR10 and Caltech256 datasets. Additional details regarding the training process can be found in \cref{subsec:details} and in supplementary \cref{sec:experimental_details}

As shown in \cref{tab:RanLI}, PEFT methods except Linear consistently demonstrate strong average results with adversarial finetuning. In particular, LoRA achieves the highest average robustness at 50.72\% for ViT, while VPT performs exceptionally well with 55.23\% on Swin.

Looking more closely at the performance on each individual dataset in \cref{fig:confounder}, we observe that adversarial linear probing achieves the strongest robustness on Caltech256. This is rather surprising, given that linear probing has the lowest computational cost compared to others and never outperforms other finetuning methods on any single dataset in standard transfer learning~\cite {vpt,glora,adaptformer}.

\subsection{Why does Adversarial Linear Probing Work?}\label{sec:why}
To understand the remarkable robustness achieved by adversarial linear probing, we then ask: \emph{what contributes to adversarial robustness on the downstream tasks given a robust pretrained model?} We hypothesize that there are two confounding factors: 1) the robustness inherited from pretrained models, and 2) the robustness achieved by adversarial finetuning. To validate our hypothesis, we conduct a comparative analysis of adversarial robustness in downstream tasks while using standard finetuning versus adversarial finetuning, given the same robust pretrained model. 

\vspace{-3mm}
\paragraph{Robustness Inherited from Pretrained Models.} 
In this section, we focus on \emph{standard finetuning} as it does not bring in additional adversarial robustness to downstream tasks. Therefore, the adversarial robustness achieved during {standard finetuning}  primarily arises from the robust pretrained model, representing the capability of each method to preserve robustness from the pretrained model. As shown in \cref{fig:confounder}, all standard finetuning methods (colored in blue) achieve non-zero robustness under PGD-10 attacks. This supports our hypothesis that robustness inherited from the pretrained model contributes to the robustness in downstream tasks. In addition, linear probing significantly outperforms all other standard finetuning methods in preserving robustness from the pretrained model. For example, linear probing exhibits significantly stronger robustness (at least 20\% higher) compared to other methods while maintaining similar accuracy on Caltech256. We explain this as linear probing avoids distorting the robust features inherited from the pretrained model by only updating the linear head. However, on CIFAR10, linear probing does not exhibit a significant advantage over robustness and displays lower clean accuracy compared to others. This is likely due to its limited capability to adapt features from the pretrained source domain to the downstream target domain.
\vspace{-3mm}
\paragraph{Robustness Achieved by Adversarial Finetuning.} By comparing adversarial robustness achieved by standard finetuning (colored in blue) and adversarial finetuning (colored in red) in \cref{fig:confounder}, we can see the extra robustness gain achieved by adversarial training during the finetuning stage. Except for Linear, adversarial finetuning contributes a significant portion of their robustness to the final robustness. For example, Full-FT on Swin, achieving a 99\% final robustness on Caltech256 through adversarial finetuning. Even VPT, which has the fewest tunable parameters after Linear, demonstrates a 78\% final robustness from adversarial finetuning. On the contrary, Linear only achieved a final robustness of 13\% from this stage, with the primal robustness originating from the pretraining model. 

Taken together, we validate our hypothesis that both robustness inherited from the pretrained model and achieved through adversarial finetuning contribute to adversarial robustness on downstream tasks. In addition, we conclude that the exceptional robustness exhibited by adversarial linear probing is mainly because linear probing excels in preserving robustness from pretrained models compared to others, which primarily obtain their robustness from adversarial training during finetuning stage. 

\subsection{Transferred Robustness Correlates with Transferred Accuracy.}
\label{subsec:transferred acc and rob}
As we have seen in \cref{fig:confounder}, adversarial linear probing achieves the highest adversarial robustness on Caltech256 but falls far behind other methods on CIFAR10. This performance discrepancy across different datasets raises a question: \emph{how can we estimate if adversarial linear probing achieves the best robustness?} Based on our analysis in \cref{sec:why}, where the robustness of adversarial linear probing primarily arises from preserving the robustness in the pretrained model, we hypothesize that the transferred robustness of adversarial linear probing is highly correlated with the transferred accuracy of standard linear probing. This suggests that if standard linear probing successfully adapts the features from the pretrained source domain to the downstream target domain, adversarial linear probing will similarly preserve the robustness in the pretrained model. 

\begin{table*}[]
\small
\centering
\setlength{\tabcolsep}{7.1pt}
\begin{tabular}{l|cccccccccc|cc}
\toprule 
\multirow{2}{*}{Methods} & \multicolumn{2}{c}{CIFAR10} 
& \multicolumn{2}{c}{CIFAR100} & \multicolumn{2}{c}{Caltech256} & \multicolumn{2}{c}{CUB200} & \multicolumn{2}{c}{Dogs} & \multicolumn{2}{|c}{Avg} \\
                         & Clean        & PGD        & Clean         & PGD        & Clean         & PGD         & Clean        & PGD       & Clean           & PGD          & Clean        & PGD       \\
                         \midrule
TWINS-AT~\cite{twins} & 91.24 & 52.73 & 70.72 & 31.08 & 76.86 & 48.40 & 68.09 & 29.24 & 64.98 & 20.58 & 74.38 & 36.41 \\
AutoLoRa~\cite{autolora} & 86.93 & 57.16 & 66.20 & 35.25 & 69.85 & 47.82 & 62.78 & 31.07 & 62.33 & 25.32 & 69.62 & 39.32 \\
\midrule
RanLI - Full-FT    & 93.02 & 59.12  & 74.36 & 35.62  & 77.35 & 49.13  & 70.83 & 31.55  & 65.85 & 24.79  & 76.28 & 40.04  \\
RanLI - Adapter    & 90.23 & 55.70  & 71.57 & 34.74  & 80.24 & 52.73  & 59.67 & 26.73  & 87.98 & 42.24  & 77.94 & 42.43  \\
RanLI - LoRA       & 90.02 & 56.88  & 71.98 & 34.83  & 77.02 & 50.25  & 64.74 & 29.65  & 84.63 & 39.24  & 77.68 & 42.17  \\
RanLI - Bias       & 90.25 & 56.94  & 71.50  & 36.72  & 79.50 & 52.43  & 66.69 & 29.88  & 87.41 & 42.75  & 79.07 & 43.74  \\
RanLI - VPT        & 90.24 & 57.38  & 72.37 & 35.60   & 79.74 & 53.07  & 61.94 & 26.46  & 86.34 & 43.10  & 78.13 & 43.12  \\
RanLI - Linear     & 84.16 & 37.30  & 67.65 & 25.46  & 81.79 & 54.47  & 69.88 & 24.56  & 91.55 & 45.45  & 79.01 & 37.45  \\
\midrule
RoLI - Full-FT     & \textbf{94.18} & \textbf{60.85}  & \textbf{76.25} & \textbf{37.38}  & \textbf{83.78} & \textbf{55.42}  & \textbf{74.28} & \textbf{32.03}  & 91.55 & 45.57  & \textbf{84.01} & \textbf{46.25}  \\
RoLI - Adapter     & 90.31 & 55.69  & 71.75 & 34.33  & 81.74 & 54.32  & 64.20 & 25.22  & \textbf{92.02} & 47.12  & 80.00 & 43.34  \\
RoLI - LoRA        & 92.23 & 58.62  & 72.87 & 35.18  & 81.79 & 54.50  & 68.88 & 26.77  & 91.20 & 46.76  & 81.39 & 44.37  \\
RoLI - Bias        & 91.83 & 58.39  & 72.17 & 35.17  & 81.78 & 54.49  & 68.14 & 25.70  & 91.52 & \textbf{47.27}  & 81.09 & 44.20  \\
\bottomrule        
\end{tabular}
\caption{\textbf{RoLI significantly improves the performance and achieves SOTA across five image classification tasks.} 
Robust Linear Initialization~(RoLI) enhances robustness by 3.88\% and clean accuracy by 2.44\% on average compared to Random Initialization (RanLI). Notably, RoLI - Full-FT achieves the best overall performance, with an accuracy of 84.01\% and 46.25\% robustness.}
\label{tab:benchmark}
\end{table*}

\begin{figure}[t]
  \centering
   \includegraphics[width=0.9\linewidth]{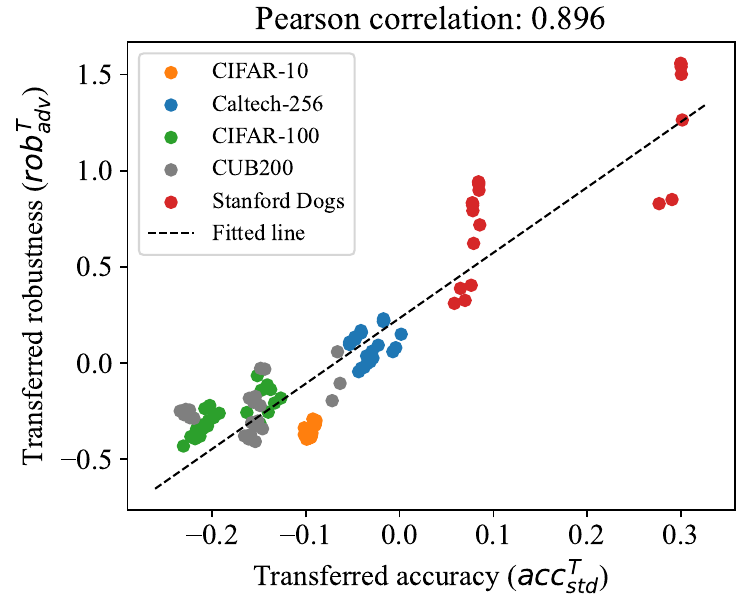}
\vspace{-3mm}
   \caption{Transferred robustness of adversarial linear probing strongly correlates with transferred accuracy of standard linear probing.}
   \label{fig:indicator}
   \vspace{-5mm}
\end{figure}

To validate this, we define \textbf{transferred accuracy} of standard linear probing ($\text{acc}^T_{std}$) and \textbf{transferred robustness} of adversarial linear probing ($\text{rob}^T_{adv}$) as follows, using full finetune as a baseline to normalize the performance across different datasets:

\begin{equation}
  \text{acc}^T_{std} = \frac{\text{acc}(LP_{std}) - \text{acc}(FT_{std})}{\text{acc}(FT_{std})}
  \label{eq:transferred accuracy}
\end{equation}
\begin{equation}
  \text{rob}^T_{adv} = \frac{\text{rob}(LP_{adv}) - \text{rob}(FT_{adv})}{\text{rob}(FT_{adv})}
  \label{eq:transferred robustness}
\end{equation}
where $\text{acc}(\cdot)$ and $\text{rob}(\cdot)$ denote the accuracy and robustness of each method, respectively. $FT_{std}$ and $LP_{std}$ represent standard full finetune and standard linear probing whereas $FT_{adv}$ and $LP_{adv}$ stand for their respective adversarial counterparts. We trained 51 models with different hyper-parameter settings across five datasets and present transferred accuracy and robustness in \cref{fig:indicator}.

In \cref{fig:indicator}, the transferred robustness of adversarial linear probing is strongly correlated with the transferred accuracy of standard linear probing, with a high Pearson correlation coefficient (0.896). When the transferred accuracy approaches or surpasses 0, the transferred robustness becomes positive, indicating the improved performance of adversarial linear probing over adversarial fully finetuning. This strong correlation validates our hypothesis that if standard linear probing successfully adapts features from the pretrained to the target domain, its adversarial counterpart is capable of effectively achieving adversarial robustness on downstream tasks.
\vspace{-1mm}
\subsection{Robust Linear Initialization (RoLI)}
\label{subsec:details}
Based on our observation that linear probing can outperform other methods when it effectively adapts features to the target domain, we propose Robust Linear Initialization (denoted as RoLI) for adversarial finetuning. Specifically, we initialize the linear head of a robust pretrained model with weights obtained through adversarial linear probing instead of random initialization. Then, we perform adversarial finetuning, using either full finetune or PEFT methods. Since adversarial linear probing excels in preserving robustness from a pretrained model, RoLI provides a more robust initialization for subsequent adversarial finetuning, maximizing the robustness inherited from a robust pretrained model. In addition, the following adversarial finetuning can further enhance robustness by introducing additional robustness through adversarial training. 
In the following sections, we will evaluate the effectiveness of RoLI applied to four different adversarial finetuning methods across five different datasets and demonstrate it achieves new state-of-the-art robustness in adversarial transfer learning.

\begin{table}[]
\vspace{-5mm}
\footnotesize
\centering
\setlength{\tabcolsep}{1.2pt}
\begin{tabular}{l|ccccc@{\hspace{1em}}c}
\toprule
Methods         & CIFAR10 & CIFAR100 & Caltech256 & CUB200 & Dogs & Avg \\

\midrule
RanLI - Full-FT & 54.9    & 31.8     & 62.6      & 49.7    & 42.5          & 48.3   \\
RanLI - Adapter & 50.8    & 29.2     & 66.7      & 40.3    & 68.5          & 51.1   \\
RanLI - LoRa    & 53.0    & 29.0     & 62.8      & 44.7    & 63.3          & 50.6   \\
RanLI - Bias    & 52.3    & 31.0     & 65.8      & 46.2    & 67.3          & 52.5   \\
RanLI - VPT     & 52.3    & 29.7     & 66.3      & 40.3    & 67.3          & 51.2   \\
RanLI - Linear  & 26.6    & 18.4     & 65.7      & 40.4    & 73.5          & 44.9   \\
\midrule
RoLI - Full-FT  & \textbf{56.3}    & \textbf{32.7}     & \textbf{70.4}      & \textbf{52.1}    & 73.5          & \textbf{57.0}   \\
RoLI - Adapter  & 51.1    & 29.1     & 65.8      & 40.6    & \textbf{75.2}          & 52.4   \\
RoLI - LoRa     & 53.9    & 30.0     & 65.7      & 44.6    & 74.3          & 53.7   \\
RoLI - Bias     & 53.9    & 29.8     & 65.7      & 43.5    & 74.8          & 53.5   \\
\bottomrule
\end{tabular}
\caption{\textbf{RoLI consistently outperforms RanLI under AutoAttack.} Note, we use ${\varepsilon} = 8/255$ attack for CIFAR10, CIFAR100, and ${\varepsilon} = 4/255$ for Caltech256, CUB200 and Stanford Dogs. RoLI shows an average 3.53 AA gain compared with RanLI, and notably, RoLI - Full-FT surpasses other methods by achieving a robustness of 57.0 under AutoAttack. The clean accuracy is illustrated in \cref{tab:benchmark}.} 
\label{tab:aa_eps4}
\vspace{-5mm}
\end{table}

\vspace{-3mm}
\paragraph{Datasets.} In our experiments, we use five datasets: CIFAR10~\cite{cifar}, CIFAR100~\cite{cifar}, Caltech256~\cite{caltech}, Caltech-UCSD Birds-200-2011 (CUB200)~\cite{cub}, and Stanford Dogs~(Dogs)~\cite{stanforddogs}. For low-resolution image datasets like CIFAR10 and CIFAR100, we resize the image to match the model input shape.  For high-resolution image datasets, we apply random resizing and cropping in training and center crop in testing. It's important to note that we integrate the resizing and normalization process directly into the model since the attacker does not have access to pre-processed images. Additional details are provided in the supplementary \cref{sec:experimental_details}.

\vspace{-3mm}
\paragraph{Methods.} We provide the results from TWINS-AT~\cite{twins} AutoLoRa~\cite{autolora} as baselines. We proceed to assess six finetuning techniques introduced in \cref{subsec:PEFTs} with random initialization. We use Swin Transformer~\cite{swin} as a backbone architecture for our experiments, whereas TWINS and AutoLoRa use ResNet50~\cite{resnet}. We apply RoLI to Full-FT, Bias, Adapter, and LoRa. We initialize the newly introduced modules in Adapter and LoRA with zero to ensure that they start with robust linear probing without distorting the robust features preserved in the pretrained model. 

\begin{table}[]
\vspace{-5mm}
\footnotesize
\centering
\setlength{\tabcolsep}{18pt}
\begin{tabular}{l|cc}
\toprule
Methods         & ${\varepsilon} = 8/255$    & ${\varepsilon} = 4/255$    \\
\midrule
RanLI - Full-FT & 20.10 & 42.50 \\
RanLI - Adapter & 36.30 & 68.50 \\
RanLI - LoRa    & 32.40 & 63.30 \\
RanLI - Bias    & 35.90 & 67.30 \\
RanLI - VPT     & 34.40 & 67.30 \\
RanLI - Linear  & 30.90 & 73.50 \\
\midrule
RoLI - Full-FT  & 30.90 & 73.50 \\
RoLI - Adapter  & 37.00 & 75.20 \\
RoLI - LoRa     & 38.40 & 74.30 \\
RoLI - Bias     & 38.20 & 74.80 \\
\bottomrule
\end{tabular}
\caption{\textbf{RoLI consistently improves adversarial robustness even against $\mathbf{\pmb{\varepsilon =} 8/255}$ AutoAttack.} RanLI - Linear performs well under ${\varepsilon} = 4/255$ AutoAttack, but poorly under ${\varepsilon} = 8/255$ AutoAttack on Stanford Dogs. This discrepancy demonstrates that adversarial linear probing obtains its robustness mainly from the pretrained model, whereas other methods obtain robustness mainly from adversarial finetuning.} 
\label{tab:aa_eps8}
\vspace{-5mm}
\end{table}

\vspace{-3mm}
\paragraph{Training Details.} 
We use the same optimizer, AdamW~\cite{adamw}, and a cosine scheduler with warm-up for all finetuning methods. To determine the optimal learning rate and weight decay values, we conduct a grid search, and the ranges of hyper-parameters along with the optimal combinations can be found in the supplementary \cref{sec:experimental_details}. We keep method-specific hyper-parameters constant across all datasets. For instance, we set the prompt length to 10 for the VPT method. To avoid adversarial overfitting~\cite{overfitadv}, we apply early stopping with the epoch that has the best performance on the validation set. During adversarial finetuning, we utilize the PGD-7 attack with ${\varepsilon} = 8/255$ and a step size of ${\alpha}= 2/255$. When reporting test accuracy and robustness, we evaluate the models under a PGD-10 attack with ${\varepsilon} = 8/255$. Additionally, we report the performance under AutoAttack~(AA), which is an ensemble attack and provides a more reliable approach to evaluate the adversarial robustness. We follow the standard AutoAttack setting, i.e., untargeted APGDCE, targeted APGD-DLR, targeted FAB~\cite{croce2020minimally}, and Square Attack~\cite{andriushchenko2020square}.

\begin{figure}[t]
\vspace{-4mm}
  \centering
   \includegraphics[width=0.85\linewidth, ,height=0.62\linewidth]{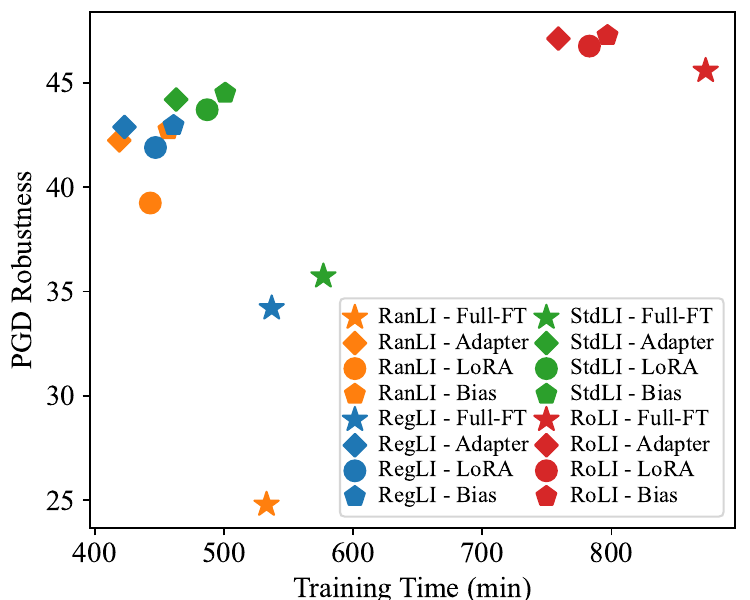}
   \vspace{-2mm}
   \caption{\textbf{Trade-off between speed and robustness.} We conduct the speed test on Stanford Dogs with random linear initialization~(RanLI), logistic regression linear initialization~(RegLI), standard linear initialization~(StdLI), and robust linear initialization~(RoLI). RoLI achieves the best performance at the cost of increased training time. Meanwhile, both RegLI and StdLI can improve their performance without incurring substantial time costs.}
   \label{fig:speed}
   \vspace{-5mm}
\end{figure}
\vspace{-3mm}
\paragraph{PGD-10 Results.} Looking at Random Initialization (RanLI) in \cref{tab:benchmark}, the same trend holds as in \cref{sec:ranli}: 1) PEFT methods except linear probing exhibit strong average results across five different datasets, with Bias achieving the best performance (79.25\% accuracy and 43.58\% robustness). 2) Linear probing achieves the highest robustness on Caltech256 and Stanford Dogs, although it falls behind others on the remaining three datasets. 

Compared to RandLI, we can observe from \cref{tab:benchmark} that RoLI significantly enhances the robustness on Caltech256 and Stanford Dogs for all finetuning methods. In particular, RoLI - Full-FT achieves a robustness improvement of 6.29\% on Caltech256 and 20.78\% on Stanford Dogs. On other datasets where adversarial linear probing with RanLI does not outperform others, RoLI performs comparable or even better than RanLI. This strongly supports that a robust initialization of the linear head is critical in adversarial transfer learning. Lastly, our RoLI - Full-FT achieves the highest performance across five datasets and sets a new state-of-the-art benchmark.

\vspace{-3mm}
\paragraph{AutoAttack Results.}
We extend our evaluation to Stanford Dogs using ${\varepsilon} = 8/255$ and ${\varepsilon} = 4/255$, as illustrated in \cref{tab:aa_eps8}. Interestingly, we observe that adversarial linear probing outperforms other methods under ${\varepsilon} = 4/255$ attack but underperforms when tested against ${\varepsilon} = 8/255$ adversarial attacks. 
We conjecture that the pretrained model lacks robustness against ${\varepsilon} = 8/255$ attacks as it is trained using ${\varepsilon} = 4/255$ adversarial attacks. 

This performance disparity confirms that adversarial linear probing inherently derives robustness from pretraining, whereas other methods predominantly achieve robustness from adversarial finetuning. Notably, RoLI consistently improves adversarial robustness even against ${\varepsilon} = 8/255$ attacks, further validating the importance of a robust linear initialization for adversarial transfer learning.

To maintain consistency with the pre-training,  we follow the conventional settings~\cite{croce2020reliable,croce2021robustbench} to set ${\varepsilon} = 8/255$ attack for low-resolution datasets~(CIFAR10, CIFAR100) and ${\varepsilon} = 4/255$ attack for high-resolution datasets~(Caltech256, CUB200, Stanford Dogs). \cref{tab:aa_eps4} demonstrates the robustness under AutoAttack and the clean accuracy is illustrated in \cref{tab:benchmark}. RoLI exhibits higher robustness over RanLI, demonstrating an average 3.53 AA gain. Notably, RoLI - Full-FT achieves a robustness of 57.0, surpassing other methods.

%% file: sec/6_ablations.tex
\section{Discussion}

\paragraph{Comparison between RoLI and StdLI.} 
Standard Linear Initialization (StdLI) refers to the use of a simple linear probing to initialize the head, while RoLI uses adversarial linear probing. In \cref{tab:StdLI}, we compare the performance of RoLI and StdLI on CIFAR-10 and Caltech-256. We observe that StdLI exhibits lower performance compared to RoLI, with a robustness decrease of 2.48 across three datasets, highlighting the effectiveness of RoLI.

\vspace{-5mm}
\paragraph{Robustness vs. Speed.}
Since robust linear initialization involves a two-step adversarial training process, it tends to be considerably slower than adversarial fine-tuning with random initialization. In this section, we show a trade-off between robustness and speed by obtaining the initialization from four different approaches: adversarial linear probing~(RoLI), standard linear probing~(StdLI), logistic regression~(RegLI), and randomization~(RanLI). From \cref{fig:speed}, it's evident that both RegLI and StdLI enhance robustness compared to random linear initialization within an acceptable time cost. In addition, RoLI achieves the highest performance but at the expense of slower speed.

\begin{table}[]
\footnotesize
\setlength\tabcolsep{0.09cm}
\centering
\begin{tabular}{l|cc|cc}
\toprule
\multirow{2}{*}{StdLI - } & \multicolumn{2}{c}{CIFAR-10} & \multicolumn{2}{|c}{Caltech-256}\\
                         & Clean      & PGD        & Clean       & PGD \\
\midrule
Full-FT & 93.34(-0.84) & 59.34(-1.51) & 83.23(-0.55) & 52.85(-2.57)\\
Bias    & 93.34(-0.04) & 58.12(-0.27) & 82.93(+1.15) & 52.93(-1.56) \\
Adapter & 90.08(-0.23) & 55.19(-0.50) & 82.72(+0.98) & 53.01(-1.31) \\
LoRa    & 90.71(-1.52) & 58.17(-0.45) & 82.87(+1.08) & 51.47(-3.03) \\
\bottomrule
\end{tabular}
\vspace{-2mm}
\caption{\textbf{StdLI is less robust than RoLI.} The values in parentheses denote the performance gap (StdLI - RoLI).}
\label{tab:StdLI}
\vspace{-5mm}
\end{table}

%% file: sec/7_conclusion_arxiv_1.tex
\section{Conclusion}

This paper systematically investigates how to achieve adversarial robustness in downstream tasks. We highlight the necessity of an adversarially robust pretraining. Given a robust pretrained model, we propose to use robust linear initialization~(RoLI) followed by adversarial full finetuning or PEFT methods to achieve the best performance. We demonstrate that RoLI outperforms random linear initialization across five image classification tasks. We hope the insights from this study greatly advance research aimed at enhancing adversarial robustness in downstream tasks.

\noindent\textbf{Acknowledgment}. We thank Chenhe Gu and Xuan Yang for the valuable discussion and insightful feedback.

%% file: sec/X_suppl_arxiv.tex
\clearpage
\setcounter{page}{1}
\maketitlesupplementary


\section*{A. Experimental Details}
\label{sec:experimental_details}

\paragraph{Pretrained Models.}
In this paper, we use ViT-B/16~\cite{vit} and Swin-B~\cite{swin} for all experiments. For standard pretrained models, we use the model in Torchvision for ImageNet-1k ViT, while obtaining the weights for CLIP ViT from hugging face \footnote{\href{https://huggingface.co/laion/CLIP-ViT-B-16-laion2B-s34B-b88K}{LAION-2B CLIP weights checkpoint}}. Additionally, we download the weights for the ImageNet-1k Swin transformer from the official implementation\footnote{\href{https://github.com/microsoft/Swin-Transformer}{Swin Transformer checkpoint}}. 

The adversarially robust models are pretrained on ImageNet-1k using PGD-3 with ${\varepsilon} = 4/255$ and a step size of ${2\varepsilon}/3$ following ~\cite{robustimagenet1k,dong2020benchmarking}. All weights for the adversarially robust pretrained models can be found at ARES 2.0 benchmark\footnote{\href{https://github.com/thu-ml/ares/tree/main/robust_training}{Adversarially robust pretrained models}}.

\paragraph{Datasets Split.}
For CIFAR10 and CIFAR100~\cite{cifar}, we follow the official train-validation-test splits. For Caltech256~\cite{caltech}, we randomly select 14,649 images (57 images per class) for training and 771 images (3 images per class) for validation, leaving the remaining images for the test set. For CUB200~\cite{cub}, we adopt the official train-test split and randomly select 600 images from the train set~(3 images per class) as the validation set. For Stanford Dogs~\cite{stanforddogs}, we also follow the official train-test split and randomly pick 1200 images from the train set~(10 images per class) as the validation set.

\paragraph{PEFT Hyperparameters.}
In this paper, we use six finetuning methods including Full Finetune, Adapter, LoRA, Bias, VPT, and Linear. For Full Finetune, Bias, and Linear, no extra hyperparameters are required as they fine-tune either part or the entirety of the model parameters.

For Adapter, we adopt the architecture proposed in~\cite{adapter2}, which involves appending the adapter module after the MLP block at each layer. Additionally, we apply a residue connection for the adapter module. Throughout the paper, we maintain a reduction factor of 8. When implementing RoLI, we initialize the downsample and upsample layers within the adapter module to zero.

We only apply the LoRA branch on the query and value projection in the self-attention block. Following~\cite{lora}, with pretrained weights $W_0$ and input $x$, the LoRA branch can be formulated as~\cref{eq:lora}
\begin{equation}
    h=W_0 x+\frac{\alpha}{r}B A x
\label{eq:lora}
\end{equation}
where $A$ and $B$ represent low-rank decomposition matrices with a rank of $r$. $\alpha$ demotes a scaling constant. We set both the rank $r$ and scaling factor $\alpha$ to 16 across all datasets. For random linear initialization~(RanLI), we adopt the initialization proposed in ~\cite{lora}, applying random Gaussian initialization for $A$ and zero initialization for $B$. For robust linear initialization~(RoLI), we zero matrices $A$ and $B$.

For VPT, we adopt VPT-Deep in ~\cite{vpt}. We set the number of tokens as 10 across all datasets. We apply uniform initialization for prompts in RanLI. We do not apply RoLI to VPT because even when prompts are zeroed, the resulting output remains distinct from the output obtained without prompts. This difference can be attributed to the influence of the softmax operation within the attention module.

\paragraph{Optimizatation Hyperparameters.}
We finetune the model for 20 epochs on CIFAR10 and CIFAR100, 40 epochs on Caltech256 and Standford Dogs, 60 epochs on CUB200. Furthermore, we conduct a grid search to determine the optimal learning rate and weight decay based on validation performance. The weight decay search range is consistent at \{0.01, 0.001, 0.0001, 0\} across all fine-tuning methods and datasets. 

The learning rate is set as $base\_lr \times b/256$. Regarding the base learning rates, for RanLI, we explore \{0.005, 0.001, 0.0001, 0.0005\} for Full Finetune and Bias, \{1.0, 0.5, 0.1, 0.05\} for Linear, \{0.5, 0.1, 0.05, 0.01\} for VPT, and \{0.05, 0.01, 0.005, 0.001\} for Adapter and LoRA. For RoLI, we expand the search range to include values an order of magnitude lower, such as \{0.005, 0.001, 0.0001, 0.0005, 0.00001, 0.00005\} for Full Finetune. 

We summarize the optimal hyperparameters combination in \cref{tab:hyper_comb}.

\paragraph{Transferred Accuracy and Transferred Robustness}
\cref{fig:indicator} demonstrates that transferred robustness is correlated with transferred accuracy. We train the model using various hyperparameters and pair the accuracy in standard linear probing with the robustness in adversarial linear probing using the same hyperparameters. For example, following the notation introduced in \cref{subsec:transferred acc and rob}, a pair obtained through finetuning is $[acc(FT_{std})_i, rob(FT_{adv})_i]$, while a pair resulting from linear probing is $[acc(LP_{std})_j, rob(LP_{adv})_j]$.

Following this, we calculated the transferred accuracy and robustness between any two pairs—comparing one from full finetuning and another from linear probing:
\begin{equation}
  (\text{acc}^T_{std})_{ij} = \frac{\text{acc}(LP_{std})_j - \text{acc}(FT_{std})_i}{\text{acc}(FT_{std})_i}
  \label{eq:transferred accuracy supp}
\end{equation}
\begin{equation}
  (\text{rob}^T_{adv})_{ij} = \frac{\text{rob}(LP_{adv})_j - \text{rob}(FT_{adv})_i}{\text{rob}(FT_{adv})_i}
  \label{eq:transferred robustness supp}
\end{equation}
Where $(\text{acc}^T_{std})_{ij}$ and $(\text{rob}^T_{adv})_{ij}$ are transferred accuracy and transferred robustness for pairs $i$ and $j$. For example, on CIFAR10, where we used 5 pairs for full fine-tuning and 6 pairs for linear probing, our plot on \cref{fig:indicator} will display 30 data points representing these comparisons.

\paragraph{Speed Analysis} In \cref{fig:speed}, we analyze the trade-off between robustness and speed. We use a single  NVIDIA RTX A6000 to conduct the experiments. For RanLI, we use the same hyperparameter search strategy and training epochs. The only difference is that RoLI applies adversarial training while RanLI does not. For RegLI, we implement logistic regression using the sklearn~\cite{sklearn} library. Additionally, we also summarize the performance for these initialization methods without further adversarial finetuning in \cref{tab:clean_trandeoff}.
\begin{table}[]
\small
\centering
\setlength{\tabcolsep}{7.1pt}
\begin{tabular}{c|ccc}
\toprule
Initialization & RegLI & StdLI & RoLI \\
\midrule
Accuracy & 92.08 & 92.75 & 91.55      \\
\bottomrule
\end{tabular}
\caption{Accuracy of different initialization methods.}
\label{tab:clean_trandeoff}
\end{table}

\begin{table*}[]
\small
\centering
\setlength{\tabcolsep}{7.1pt}
\begin{tabular}{l|cccccccccc}
\toprule
Datasets & \multicolumn{2}{c}{CIFAR10} & \multicolumn{2}{c}{CIFAR100} & \multicolumn{2}{c}{Caltech256} & \multicolumn{2}{c}{CUB200} & \multicolumn{2}{c}{Dogs} \\
Hyperparameters & Base Lr     & Wd     & Base Lr     & Wd     & Base Lr      & Wd     & Base Lr     & Wd     & Base Lr      & Wd     \\
\midrule
RanLI - Full-FT & 0.0005 & 0.01   & 0.0005 & 0.001  & 0.0005  & 0.0001 & 0.0005 & 0.001  & 0.0005  & 0.01   \\
RanLI - Adapter & 0.05   & 0.01   & 0.005  & 0.0001 & 0.005   & 0.0001 & 0.05   & 0.0001 & 0.005   & 0.0001 \\
RanLI - LoRa    & 0.01   & 0.0001 & 0.05    & 0      & 0.001   & 0.0001 & 0.005  & 0.0001 & 0.005   & 0.0001 \\
RanLI - Bias    & 0.005  & 0.001  & 0.005  & 0.001  & 0.001   & 0.01   & 0.005  & 0.001  & 0.001   & 0.0001 \\
RanLI - VPT     & 0.5    & 0.001  & 0.5    & 0.01   & 0.05    & 0.001  & 0.05   & 0.001  & 0.05    & 0.0001 \\
RanLI - Linear  & 0.05   & 0.01   & 0.1    & 0.001  & 0.5     & 0.01   & 1      & 0.001  & 1       & 0.01   \\
\midrule
RoLI - Full-FT  & 0.0001 & 0.01   & 0.0001 & 0.001  & 0.0001  & 0.0001 & 0.0001 & 0.01   & 0.00005 & 0.001  \\
RoLI - Adapter  & 0.05   & 0.01   & 0.05   & 0.0001 & 0.0005  & 0      & 0.05   & 0.001  & 0.0001  & 0.0001 \\
RoLI - LoRa     & 0.001  & 0.0001 & 0.001  & 0.0001 & 0.0001  & 0      & 0.005  & 0.0001 & 0.001   & 0.01   \\
RoLI - Bias     & 0.005  & 0.001  & 0.005  & 0.01   & 0.00001 & 0      & 0.0005 & 0.0001 & 0.00005 & 0.0001 \\
\bottomrule
\end{tabular}
\vspace{-2mm}
\caption{Summary of the optimal hyperparameter combination across five datasets.}
\vspace{-2mm}
\label{tab:hyper_comb}
\end{table*}

\begin{table*}[]
\small
\centering
\setlength{\tabcolsep}{7.1pt}
\begin{tabular}{c|c|cccccc|cccccc}
\toprule
\multirow{3}{*}{Model} & \multirow{3}{*}{Methods} & \multicolumn{6}{c|}{CIFAR10}                                                      & \multicolumn{6}{c}{Caltech256}                                                   \\
                       &                          & \multicolumn{2}{c}{IN} & \multicolumn{2}{c}{CLIP} & \multicolumn{2}{c|}{Robust IN} & \multicolumn{2}{c}{IN} & \multicolumn{2}{c}{CLIP} & \multicolumn{2}{c}{Robust IN} \\
                       &                          & Clean      & PGD       & Clean       & PGD        & Clean         & PGD           & Clean      & PGD       & Clean       & PGD        & Clean         & PGD           \\
\midrule
\multirow{6}{*}{ViT}   & Full-FT                  & 87.91      & 50.45     & 78.53       & 45.26      & 89.32         & 51.62         & 52.82      & 25.91     & 37.16       & 5.29       & 71.01         & 36.68         \\
                       & Adapter                  & 64.19      & 29.13     & 58.70       & 28.51      & 88.17         & 52.63         & 46.24      & 10.94     & 7.95        & 6.60        & 78.11         & 46.51         \\
                       & LoRa                     & 82.42      & 47.68     & 69.43       & 41.01      & 90.03         & 56.28         & 1.01       & 1.07      & 4.86        & 4.86       & 75.22         & 45.16         \\
                       & Bias                     & 56.30       & 31.19     & 55.85       & 32.06      & 88.29         & 52.94         & 41.07      & 16.07     & 8.59        & 4.95       & 77.09         & 47.19         \\
                       & VPT                      & 68.61      & 31.4      & 40.43       & 24.87      & 88.64         & 54.25         & 36.94      & 9.89      & 7.70         & 4.86       & 75.54         & 46.79         \\
                       & Linear                   & 52.99      & 0.21      & 74.88       & 0.00       & 84.42         & 35.81         & 53.80      & 0.50      & 26.06       & 0.54       & 79.35         & 48.22         \\
\midrule
\multirow{6}{*}{Swin}  & Full-FT                  & 87.54      & 51.79     & -           & -          & 93.02         & 59.12         & 62.71      & 34.77     & -           & -          & 77.35         & 49.13         \\
                       & Adapter                  & 70.32      & 32.83     & -           & -          & 90.23         & 55.70         & 41.48      & 10.27     & -           & -          & 80.24         & 52.73         \\
                       & LoRa                     & 80.55      & 47.67     & -           & -          & 90.02         & 56.88         & 0.29       & 0.28      & -           & -          & 77.02         & 50.25        \\
                       & Bias                     & 62.25      & 35.02     & -           & -          & 90.25         & 56.94         & 1.44       & 1.44      & -           & -          & 79.50         & 52.43         \\
                       & VPT                      & 66.90      & 35.12     & -           & -          & 90.24         & 57.38         & 45.22      & 18.36     & -           & -          & 79.74         & 53.07         \\
                       & Linear                   & 42.13      & 1.62      & -           & -          & 84.16         & 37.30         & 25.92      & 0.60      & -           & -          & 81.79         & 54.47         \\
\bottomrule
\end{tabular}
\vspace{-2mm}
\caption{
\textbf{Numerical results for \cref{fig:pretrain}.} We omit the results obtained from CLIP with Swin Transformer because the pretrained model is unavailable.
}
\vspace{-2mm}
\label{tab:numerical_pretrain}
\end{table*}

\begin{table*}[]
\small
\centering
\setlength{\tabcolsep}{7.1pt}
\begin{tabular}{c|c|ccccc|ccccc}
\toprule
\multirow{3}{*}{Model} & \multirow{3}{*}{Methods} & \multicolumn{5}{c|}{CIFAR10}                                                                  & \multicolumn{5}{c}{Caltech256}                                                               \\
                       &                          & \multicolumn{2}{c}{Standard FT} & \multicolumn{2}{c}{Adversarial FT} & \multirow{2}{*}{$\Delta$ PGD} & \multicolumn{2}{c}{Standard FT} & \multicolumn{2}{c}{Adversarial FT} & \multirow{2}{*}{$\Delta$ PGD} \\
                       &                          & Clean          & PGD            & Clean            & PGD             &                        & Clean          & PGD            & Clean            & PGD             &                        \\
\midrule
\multirow{6}{*}{ViT}   & Full-FT                  & 97.58          & 9.57           & 89.32            & 51.62           & 42.05                  & 80.02          & 6.41           & 71.01            & 36.68           & 30.27                  \\
                       & Adapter                  & 97.65          & 8.15           & 88.17            & 52.63           & 44.48                  & 85.59          & 9.65           & 78.11            & 46.51           & 36.86                  \\
                       & LoRa                     & 97.80           & 21.58          & 90.03            & 56.28           & 34.70                  & 84.85          & 6.49           & 75.22            & 45.16           & 38.67                  \\
                       & Bias                     & 97.95          & 9.35           & 88.29            & 52.94           & 43.59                  & 86.66          & 4.09           & 77.09            & 47.19           & 43.10                   \\
                       & VPT                      & 97.51          & 10.22          & 88.64            & 54.25           & 44.03                  & 85.27          & 8.96           & 75.54            & 46.79           & 37.83                  \\
                       & Linear                   & 89.57          & 19.87          & 84.42            & 35.81           & 15.94                  & 82.85          & 32.72          & 79.35            & 48.22           & 15.50                   \\
\midrule
\multirow{6}{*}{Swin}  & Full-FT                  & 98.66          & 0.92           & 93.02            & 59.12           & 58.20                  & 87.89          & 0.16           & 77.35            & 49.13           & 48.97                  \\
                       & Adapter                  & 98.68          & 4.26           & 90.23            & 55.70           & 51.44                  & 87.35          & 3.90           & 80.24            & 52.73           & 48.83                  \\
                       & LoRa                     & 98.49          & 13.51          & 90.02            & 56.88           & 43.37                  & 87.24          & 1.49           & 77.02            & 50.25           & 48.76                 \\
                       & Bias                     & 98.80          & 7.06           & 90.25            & 56.94           & 49.88                  & 88.49          & 0.79           & 79.50            & 52.43           & 51.64                  \\
                       & VPT                      & 98.39          & 14.62          & 90.24            & 57.38           & 42.76                  & 86.61          & 11.50          & 79.74            & 53.07           & 41.57                  \\
                       & Linear                   & 90.33          & 21.91          & 84.16            & 37.30           & 15.39                  & 85.01          & 47.10          & 81.79            & 54.47           & 7.37                   \\
\bottomrule
\end{tabular}
\caption{Numerical results for \cref{fig:confounder} and \cref{tab:RanLI}.}
\label{tab:numerical_confounder}
\end{table*}

\section*{B. More Ablation Studies}
\paragraph{RoLI Mitigates Overfitting.}

\begin{figure}[t]
\vspace{-4mm}
  \centering
   \includegraphics[width=0.9\linewidth]{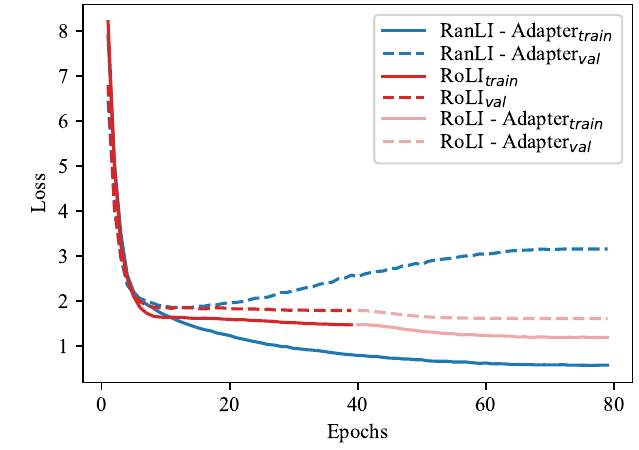}
      
    \vspace{-1mm}
   \caption{\textbf{RoLI mitigates overfitting.} For RoLI, the loss for the adversarial linear probing stage and Adapter finetuning stage is denoted as red and light red lines, respectively. The decreasing validation loss demonstrates that RoLI can mitigate overfitting.}
   \label{fig:overfit}
\end{figure}

Overfitting is a general phenomenon for adversarial training, as highlighted in prior research~\cite{overfitadv, understandoverfitting}. 
Although intuitively PEFT methods could mitigate overfitting as they only update a small number of parameters, we observe that they still suffer from overfitting.  Taking adversarial Adapter on Stanford Dogs as an example in \cref{fig:overfit}, RanLI suffers from overfitting, while RoLI effectively mitigates this phenomenon. 
Specifically, adversarial linear probing (RoLI) avoids overfitting possibly because it does not change the features during finetuning. In addition, when we initialize adversarial Adapter with RoLI, we can use a smaller learning rate during finetuning, resulting in a further decrease in loss without overfitting.

\begin{figure}[t]
  \centering
   \includegraphics[width=1\linewidth]{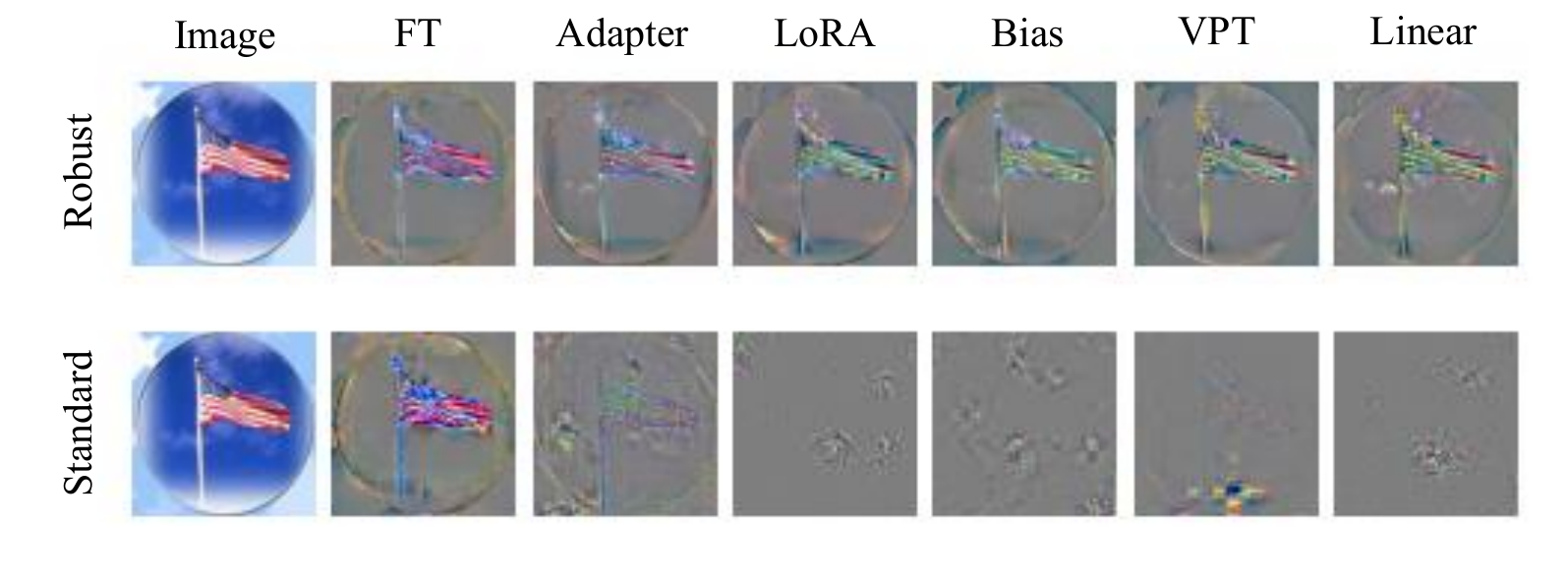}
   
   \caption{A robust pretrained model exhibits more semantically structured gradients compared to a standard pretrained model.}
   \label{fig:gradient}
   \vspace{-5mm}
\end{figure}

\paragraph{Image Gradient Visualization.}

In this section, we present visualizations of the loss gradient with respect to the input image for different PEFT methods using either an adversarially robust or non-robust pretrained model on the Caltech256 dataset. \cref{fig:gradient} illustrates semantically structured gradients obtained from adversarially robust pretraining, whereas non-robust pretraining yields less semantically structured gradients, except fully finetuning. This observation aligns with our results, where fully finetuning outperforms other methods when a non-robust pretraining is applied. In contrast to fully finetuning, the image gradients from PEFT methods contain less semantic information, further validating our results.

\section*{C. RoLI Pseudo Code}
\vspace{-4mm}
\begin{algorithm}
    \footnotesize
    \caption{RoLI with adversarial finetuning}
    \begin{algorithmic}
        \STATE\textbf{Input:} Dataset $D$, Model $\theta$,
        Parameters in linear head $\theta_{head}$, Tunable parameters in each adversarial finetuning method $\theta_{ft}$, Perturbation bound $\varepsilon$, Loss function $\mathcal{L}$ \newline
        \textbf{Output:} Adversarially robust model $\theta^*$ \newline
        // Initialize tunable parameters in linear head. \newline
        Random initialize $\theta_{head}$ \newline
        // Robust Linear Initialization \newline
        $\hat{\theta} \gets \theta_{head}$, $\theta_{frozen} \gets \theta \setminus \theta_{head}$ \newline
        $\hat{\theta} \gets \min _{\hat{\theta}} \mathbb{E}_{(x, y) \sim D}\left[\max _{\|\delta\|_\infty \leq \varepsilon} \mathcal{L}(x+\delta, y ; \hat{\theta} \cup \theta_{frozen})\right]$ \newline
        $\theta \gets \hat{\theta} \cup \theta_{frozen}$ \newline
        // Adversarial FT, tunable parameters shown in Fig. 1 \newline
        $\hat{\theta} \gets \theta_{ft}$, $\theta_{frozen} \gets \theta \setminus \theta_{ft}$ \newline
        $\hat{\theta} \gets \min _{\hat{\theta}} \mathbb{E}_{(x, y) \sim D}\left[\max _{\|\delta\|_\infty \leq \varepsilon} \mathcal{L}(x+\delta, y ; \hat{\theta} \cup \theta_{frozen})\right]$ \newline
        $\theta^* \gets \hat{\theta} \cup \theta_{frozen}$
    \end{algorithmic}
\label{algo}
\end{algorithm}

\vspace{-4mm}
\section*{D. Numerical Results}
For \cref{fig:intro}, the numerical results are in \cref{tab:benchmark} average column. The results for TWINS-AT~\cite{twins} and AutoLoRa~\cite{autolora} are from their original papers. We show the numerical results for \cref{fig:pretrain} in \cref{tab:numerical_pretrain}. We do not include the CLIP results for Swin Transformer because the CLIP pretrained weights are only available for ResNet-50~\cite{resnet} and ViT~\cite{vit}. \cref{tab:numerical_confounder} displays the numerical results corresponding to \cref{fig:confounder} and \cref{tab:RanLI}.